\documentclass[10pt,twocolumn,letterpaper]{article}

\usepackage{cvpr}
\usepackage{times}
\usepackage{epsfig}
\usepackage{graphicx}
\usepackage{amsmath}
\usepackage{amssymb}

\usepackage{enumitem}
\usepackage{multirow}
\usepackage{soul}
\usepackage{caption}
\usepackage{subcaption}

\usepackage[pagebackref=true,breaklinks=true,letterpaper=true,colorlinks,bookmarks=false]{hyperref}

\cvprfinalcopy 

\DeclareMathOperator*{\argmax}{arg\,max}

\ifcvprfinal\fi
\begin{document}

\title{Distilling Knowledge from Refinement in Multiple Instance Detection Networks}

\author{Luis Felipe Zeni \\
{\tt\small luis.zeni@inf.ufrgs.br }
\and
Claudio R. Jung \\
{\tt\small crjung@inf.ufrgs.br }
\and
\hfill \break
Institute of Informatics, Federal University of Rio Grande do Sul, Brazil
}

\maketitle

\begin{abstract}
Weakly supervised object detection (WSOD) aims to tackle the object detection problem using only labeled image categories as supervision. A common approach used in WSOD to deal with the lack of localization information is Multiple Instance Learning, and in recent years methods started adopting Multiple Instance Detection Networks (MIDN), which allows training in an end-to-end fashion. In general, these methods work by selecting the best instance from a pool of candidates and then aggregating other instances based on similarity. In this work, we claim that carefully selecting the aggregation criteria can considerably improve the accuracy of the learned detector. We start by proposing an additional refinement step to an existing approach (OICR), which we call refinement knowledge distillation. Then, we present an adaptive supervision aggregation function that dynamically changes the aggregation criteria for selecting boxes related to one of the ground-truth classes, background, or even ignored during the generation of each refinement module supervision. Experiments in Pascal VOC 2007 demonstrate that our Knowledge Distillation and smooth aggregation function significantly improves the performance of OICR in the weakly supervised object detection and weakly supervised object localization tasks. These improvements make the Boosted-OICR competitive again versus other state-of-the-art approaches.

\end{abstract}

\section{Introduction}

Supervised object detection has been achieving increasingly better results in terms of accuracy and speed along the past years~\cite{redmon2017yolo9000, liu2016ssd}. The main drawback of these methods is the need for annotated bounding boxes, which is a tedious, error-prone, time-consuming, and expensive task. 
The annotation cost directly impacts the viability of deployment of these detectors in real-world applications, particularly when starting from scratch for a specific application. One approach that researchers are exploring to alleviate the annotation cost is Weakly Supervised Object Detection (WSOD), where the object detector is trained using only image category annotations (presence or absence of interest classes in the image), which is much easier and faster to generate.

Most WSOD methods~\cite{Cinbis2016, Bilen2016, Tang2017_OIRC, Diba2017, Tang2018_PCL, Wan2019_CMIL} follow the Multiple Instance Learning (MIL) pipeline~\cite{Dietterich97} to train detectors using only image category level annotations. 
In the adaptation of MIL to the WSOD task, each image is considered a bag of positive and negative object proposals generated by object proposal methods such as Selective Search~\cite{Uijlings2013} or Edge Boxes~\cite{Zitnick2014}. 
The training process in the MIL framework encompasses two steps: (i) to train an instance selector to compute the object score of each object proposal; (ii)  to select the proposal with the highest score and use it to mine positive instances and train detector estimators. 
The majority of recent methods explore features extracted by Convolutional Neural Networks (CNN)  as an off-the-shelf feature extractor~\cite{Cinbis2016, Li2016} or train an end-to-end Multiple Instance Detection Network (MIDN)~\cite{Bilen2016}.

The lack of localization supervision during the training process, as expected, makes detection accuracy of WSOD methods worse than its supervised counterparts.
However, the promise of a lower annotation cost attracted the efforts of many researchers to WSOD, and significant improvements were achieved in recent years exploring a variety of strategies~\cite{Cinbis2016,Tang2017_OIRC, Tang2018_RPN, Tang2018_PCL, Wan2019_CMIL}. 


In this paper, we focused on the instance mining step of MIL-based methods, and used a modification of an existing baseline approach as a proof-of-concept. More precisely, we propose improvements to boost the performance of OICR, which we call Boosted-OICR (BOICR). We first observed that it is possible to extract extra information from the refinement modules to boost the detection mAP of OICR, which we call refinement knowledge distillation. We also propose an adaptive supervision aggregation function that dynamically changes the IoU threshold to select boxes that will be aggregated as belonging to one of the ground-truth class, background, or ignored during the generation of each refinement module supervision. The selection process follows the principle that at the beginning of the training is better to aggregate boxes with small IoU (since the best instance is typically small and comprehends a small portion of the object, such as the face for a person or cat).  To avoid an overgrowth of the object-related proposals, the IoU threshold is tightened as the training phase advances. We also embedded an adapted version of the ``trick'' proposed in~\cite{Tang2018_RPN}, which ignores boxes with small intersection in the refinement losses. We evaluate our method in Pascal VOC 2007, and our approach presents competitive state-of-art results both in detection mAP and CorLoc mAP.

Our main contributions in this paper are the introduction of: i) a module to distill extra knowledge from refinement agents; and ii) an adaptive supervision aggregation function to mine candidate instances. Next, we present the state-of-the-art on WSOD, and then describe the proposed methodology with the experimental results and conclusions.

\begin{figure*}[t]
\begin{center}
  \includegraphics[width=0.90\linewidth]{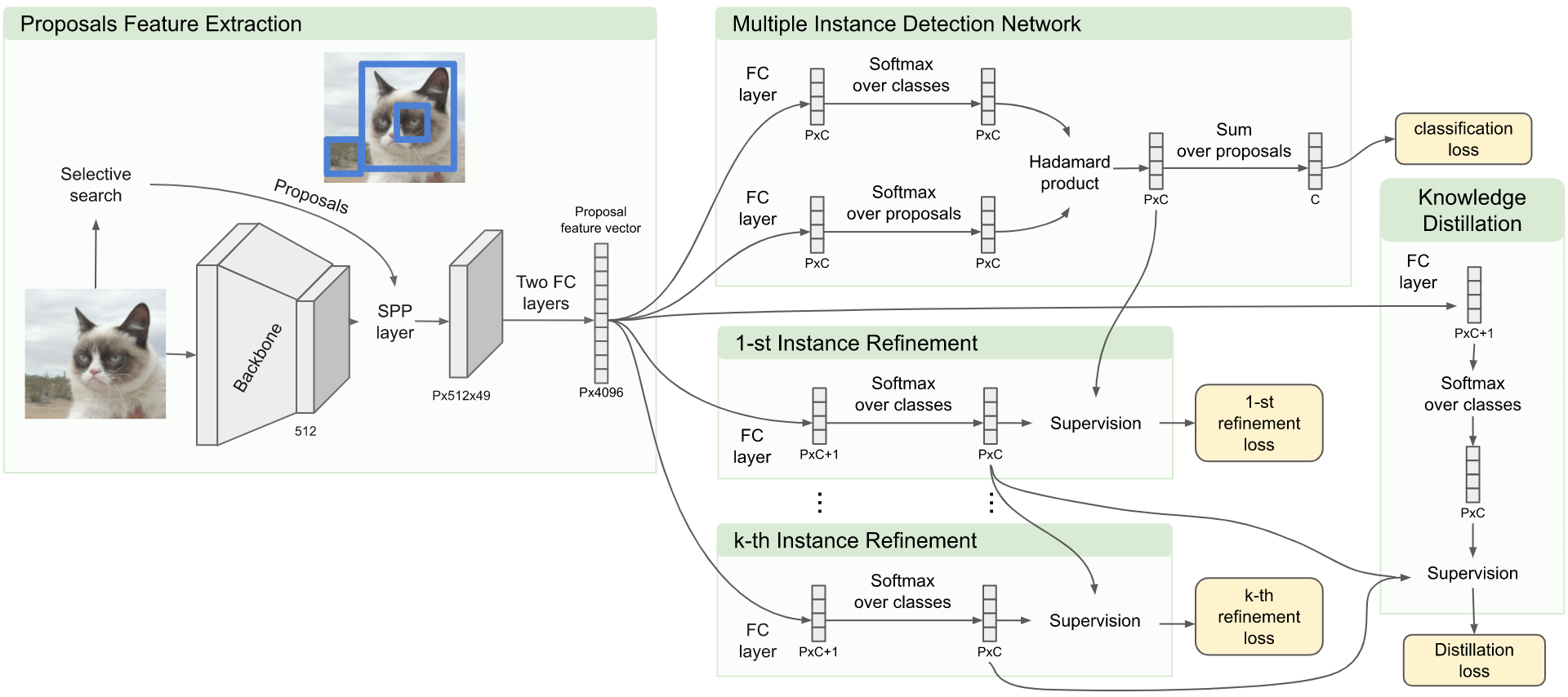}
\end{center}
\caption{
    The proposed architecture and its four modules. The proposals feature extraction module uses an SSP layer to extract features from proposals generated by selective search. The multiple instance detection network module learns to select the best proposal instance and generates an image classification score. The instance refinement modules have $k$ instances, and each one learns to refine instances from its predecessor result. Finally, the knowledge distillation module aggregates all the knowledge learned by all the $K$ refinement agents. 
}
\label{fig:architeture_diagram}
\end{figure*}

\section{Related Work}

There is a considerable number of WSOD works that precede the CNN era~\cite{russakovsky2012object, siva2013looking, song2014weakly}. However, we focus on CNN- based methods as all state-of-the-art methods rely on CNN architectures. The adoption of CNN features was not immediate, and initial works started combining the CNN features with features extracted by other kinds of feature descriptors. Cinbis et. al.~\cite{Cinbis2016} proposed a multi-fold multiple instance learning training procedure, which splits the positive instances in $K$ training folds. The method combines the Fisher Vector with CNN features as descriptors, and an objectness refinement is proposed to improve localization accuracy. Since a pre-trained CNN is only used as a feature extractor, its weights are not fine-tuned, which can lead to lower accuracy. Li et al.~\cite{Li2016} introduced a two-stage adaptation algorithm. The first stage fine-tunes the network to collect class-specific object proposals with higher precision; the second uses confident object candidates to optimize the CNN representations to turn image classifiers into object detectors gradually. 
A drawback of the method is the need for individually forwarding each region proposal into CNN to extract features, making the whole process very slow. This problem is solved in more recent methods using Spatial Pyramidal Pooling (SPP)~\cite{he2015spatial}. 

Bilen et al.~\cite{Bilen2016} proposed a two-stream method, where one stream performs classification and the other detection. The output of both streams is combined into a global scoring matrix by taking the Hadamard product of the two streams. The classification scores are calculated by summing the values in the proposals dimension of this matrix. Tang et al.~\cite{Tang2017_OIRC} improved the smoothed version of MIL proposed by~\cite{Bilen2016} using an online instance classification refinement that utilizes cascaded refinement modules to increase the detection performance, where each refinement steep makes the detector able to detect larger objects parts during training gradually. In~\cite{Tang2018_PCL}, the refinement process of~\cite{Tang2017_OIRC} is further improved, adding proposal clusters to select one or more supervision boxes during the training. Selecting more than one supervision box is interesting because, usually, objects can have multiple parts and also have multiple instances present in the image. However, a limitation of the clustering process is that it increases the computational cost making the whole training process slower. Our Boosted-OICR has a better mAP result than~\cite{Tang2018_PCL} without using the clustering process.

%

Diba et. al.~\cite{Diba2017} proposed a three-stage cascaded method that mines boxes from Class Activation Maps (CAM). The first stage is inspired by~\cite{zhou2016learning}, which uses a fully convolutional CNN with global average pooling (GAP) to create the CAMs in conjunction with the classification scores. The second stage uses the CAM from the first stage as supervision to generate a segmentation map that is used to select a set of candidate bounding boxes using the connective algorithm from~\cite{zhou2016learning}. Finally, the features of the candidate boxes are extracted by an SPP layer~\cite{he2015spatial}, and a MIL algorithm is applied to select the best candidate boxes for each class. In the same direction, Wei et al.~\cite{Wei2018} introduced a method that uses CAMs to mine tight object boxes by exploiting segmentation confidence maps. The segmentation confidence maps are employed to evaluate the objectness scores of proposals according to two properties -- purity and completeness --, and the detection process is based on~\cite{Tang2017_OIRC}. Although the idea of using CAMs to guide the selection of the supervision boxes is interesting, the training process of \cite{Diba2017,Wei2018} is overly complex.

Wan et al.~\cite{Wan2019_MEL} proposed a min-entropy latent model to measure the randomness of object localization. The learning process operates with two network branches. The first branch is designated for discovering objects using a global min-entropy layer that defines the distribution of object probability. This discovery process targets at finding candidate object cliques, which is a proposal with high object confidence. The second branch is designated to localize objects using a local min-entropy layer and a softmax layer. The local min-entropy layer classifies the object candidates in a clique into pseudo objects and hard negatives by optimizing the local entropy.

Non-convexity is also a common problem in multiple instance learning, which might lead to sub-optimal results.  Wan et al. ~\cite{Wan2019_CMIL} introduced a continuation optimization method that uses a series of smoothed loss functions to approximate the target (desired) loss, claiming that this smoothed process alleviates the non-convexity problem in MIL. The authors also propose a parametric strategy, for instance, subset partition, which is combined with a deep neural network to activate a full object extent. In contrast, Tang et al.~\cite{Tang2018_RPN} proposed a two-stage region proposal network that explores the responses in mid-layers of a network to create object proposals. The process creates coarse proposals using an objectness score metric and sliding window boxes. Later, the coarse proposals are refined proposals using a region-based CNN classifier, which are used to train the network proposed in \cite{Tang2017_OIRC}. 

In summary, existing WSOD approaches vary regarding the selection of candidate proposals, the strategy for mining instances, and the underlying classification network that guides the supervision, which leads to different levels of complexity for both implementation and training times. 
This paper focuses mostly on the instance selection part, and we used the continuation function proposed in \cite{Wan2019_CMIL} as inspiration to adaptively select positive and negative instances. We also present and additional step to the refinement supervision of~\cite{Tang2017_OIRC}.
The proposed method is presented next.

\section{The Proposed Approach}
\label{sec:method}

Since we propose improvements to boost OICR's pipeline~\cite{Tang2017_OIRC}, we will try to follow the same notation of the original paper, and Fig.~\ref{fig:architeture_diagram} shows a high-level diagram of all stages of the proposed architecture.
The first stage aims to extract feature vectors from a given image, and candidate proposals are extracted using selective search~\cite{Uijlings2013}. The image and the extracted proposals feed a CNN backbone with SPP to produce a fixed-size feature map to each proposal. The proposals feature maps are converted to proposal feature vectors using two fully connected (fc) layers, which  are branched into three different stages.
The two first stages are similar to~\cite{Tang2017_OIRC} stages, where the first one trains a basic instance classifier, and the second stage trains a set of $K$ refinement agents. The $k^{th}$ refinement agent uses as supervision the output from the previous agent $\{k-1\}$, and the supervision for the $1^{st}$ refinement agent ($k=1$) comes from the instance classifier branch. The third state, proposed by us, utilizes the knowledge of all $K$ refinement agents to train a new agent. We call this process knowledge distillation as it aims to extract extra knowledge during the refinement process.  

In this section, we will explain all the employed stages in detail. Also, in section \ref{smooth_supervision}, we explain the adaptive supervision aggregation function that is employed by all refinement agents during the learning process.

\subsection{Instance selection}

Following~\cite{Tang2017_OIRC}, we use the method proposed by~\cite{Bilen2016} because of its effectiveness and implementation convenience. 
The instance selection works by branching the proposal feature vectors into two streams, and each stream starts with an fc layer to produce two matrices $\mathbf{x}^c$, $\mathbf{x}^d \in \mathbb{R}^{C\times|R|}$, where $C$ is the number of classes and $|R|$ is the number of proposals. 
A softmax function is applied to both matrices along different dimensions, yielding 
\begin{equation}
\sigma(\mathbf{x}^c)]_{ij}= \frac{e^{x^c_{ij}}}{\sum^{C}_{k=1}e^{x^c_{kj}}}, ~~
\sigma(\mathbf{x}^d)]_{ij}= \frac{e^{x^d_{ij}}}{\sum^{|R|}_{k=1}e^{x^c_{ik}}}.
\end{equation}

The two streams are then combined to generate proposal scores using Hadamard (element-wise) matrix product, yielding $\mathbf{x}^R = \sigma(\mathbf{x}^c) \odot  \sigma(\mathbf{x}^d)$. Finally, the classification score
$\phi_c \in (0,1)$ for class $c$ is obtained by by summing over proposal dimensions, i.e,. $\phi_c = \sum^{|R|}_{r=1} \mathbf{x}^R_{cr}$. We train the instance classifier using multi-class cross entropy loss, defined as
\begin{equation}
\label{eq:loss_cls}
L_{class} = -\sum^C_{c=1}  y_c \log \phi_c + (1 - y_c)  \log ( 1 - \phi_c),
\end{equation} 
where  $y_c = \in {0,1}$ indicates if the image contains any instance of class $c$ in the image. More details can be found in~\cite{Bilen2016, Tang2017_OIRC}

\begin{figure}
\centering
\begin{subfigure}{.4\linewidth}
    \centering
    \includegraphics[width=0.99\linewidth]{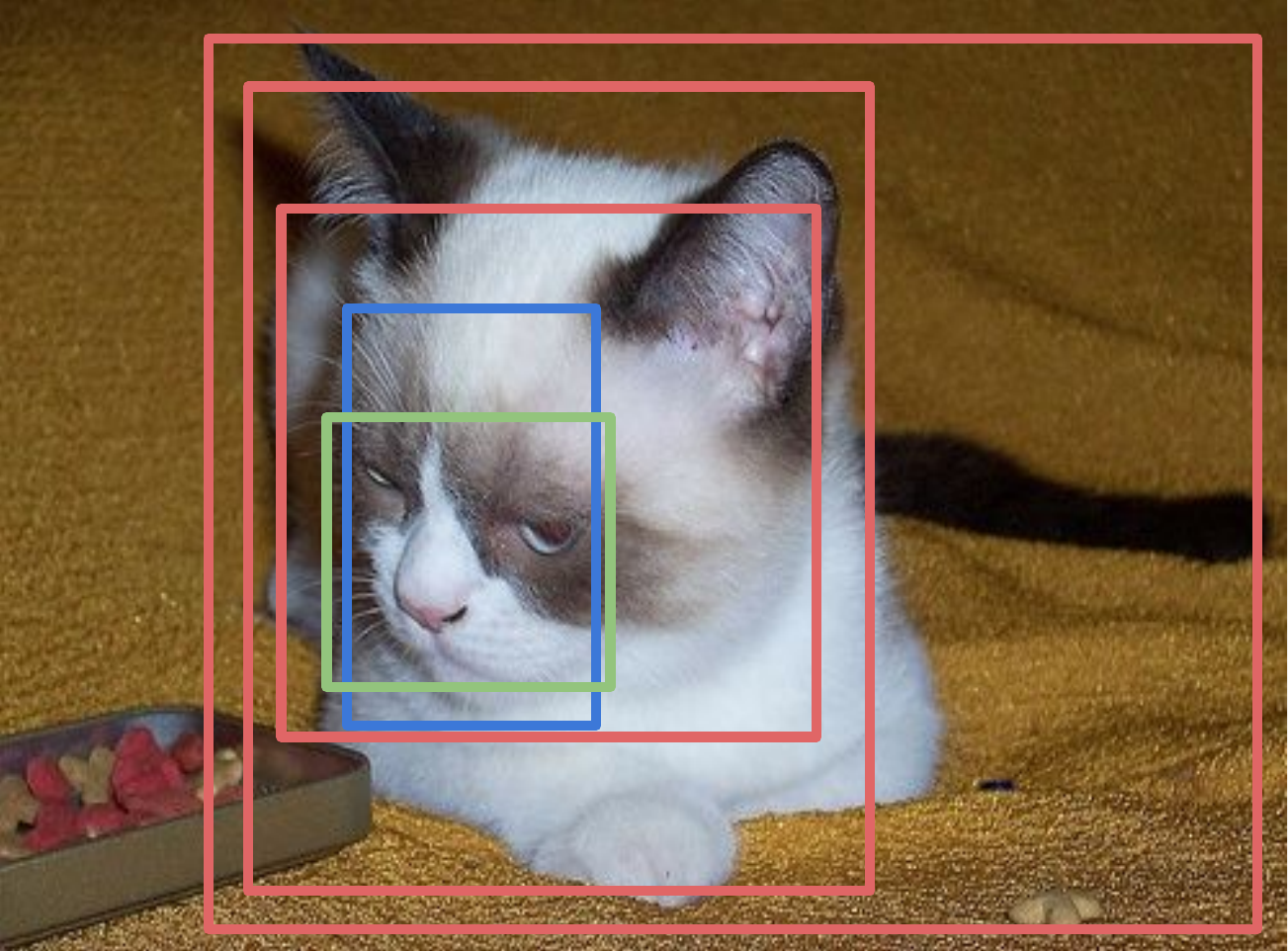}
    \caption*{$\lambda=0.5$}
\end{subfigure}%
\begin{subfigure}{.4\linewidth}
    \centering
    \includegraphics[width=0.99\linewidth]{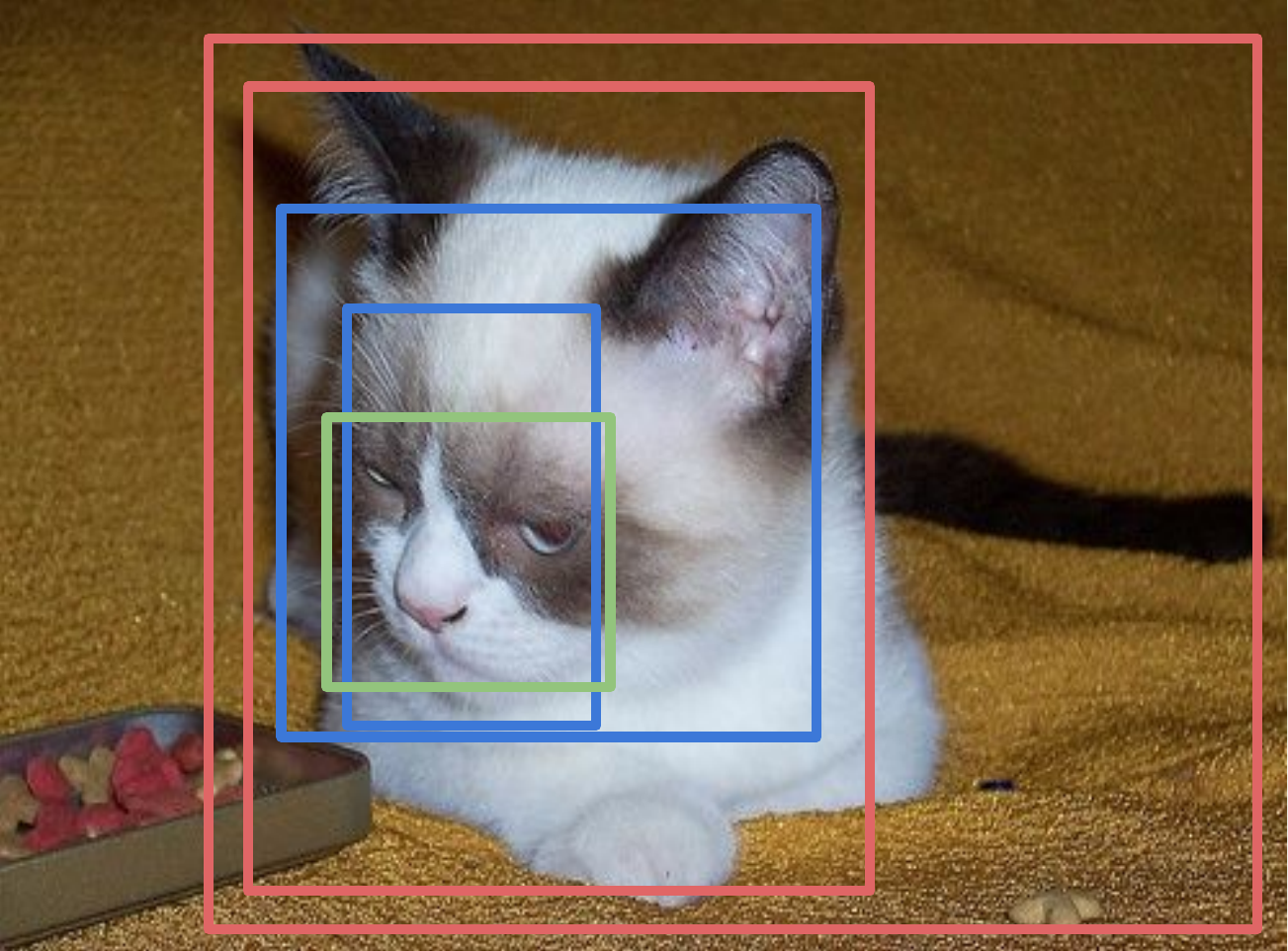}
    \caption*{$\lambda=0.25$}
    
\end{subfigure}
\begin{subfigure}{.4\linewidth}
    \centering
    \includegraphics[width=0.99\linewidth]{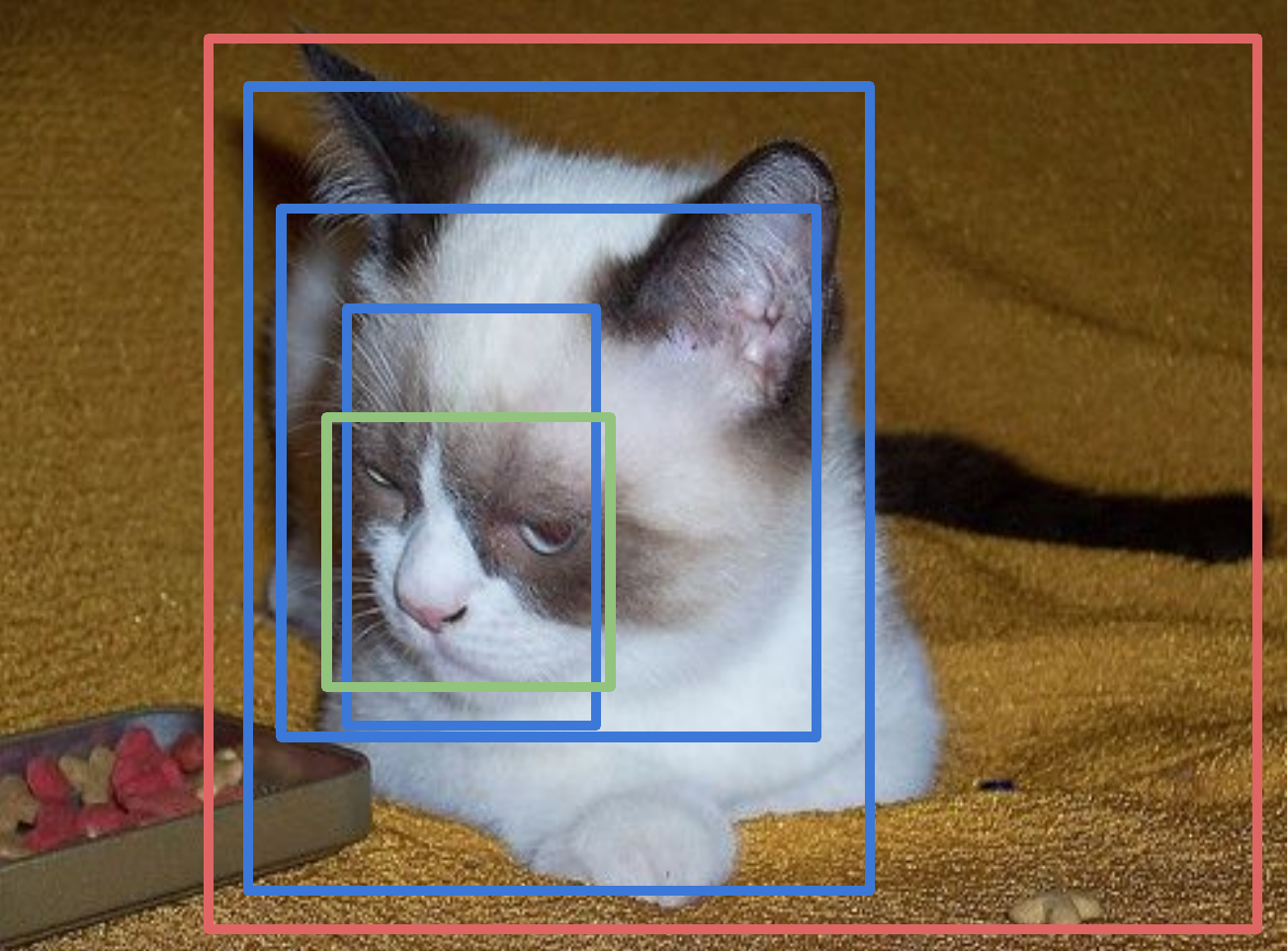}
    \caption*{$\lambda=0.1$}
\end{subfigure}%
\begin{subfigure}{.4\linewidth}
    \centering
    \includegraphics[width=0.99\linewidth]{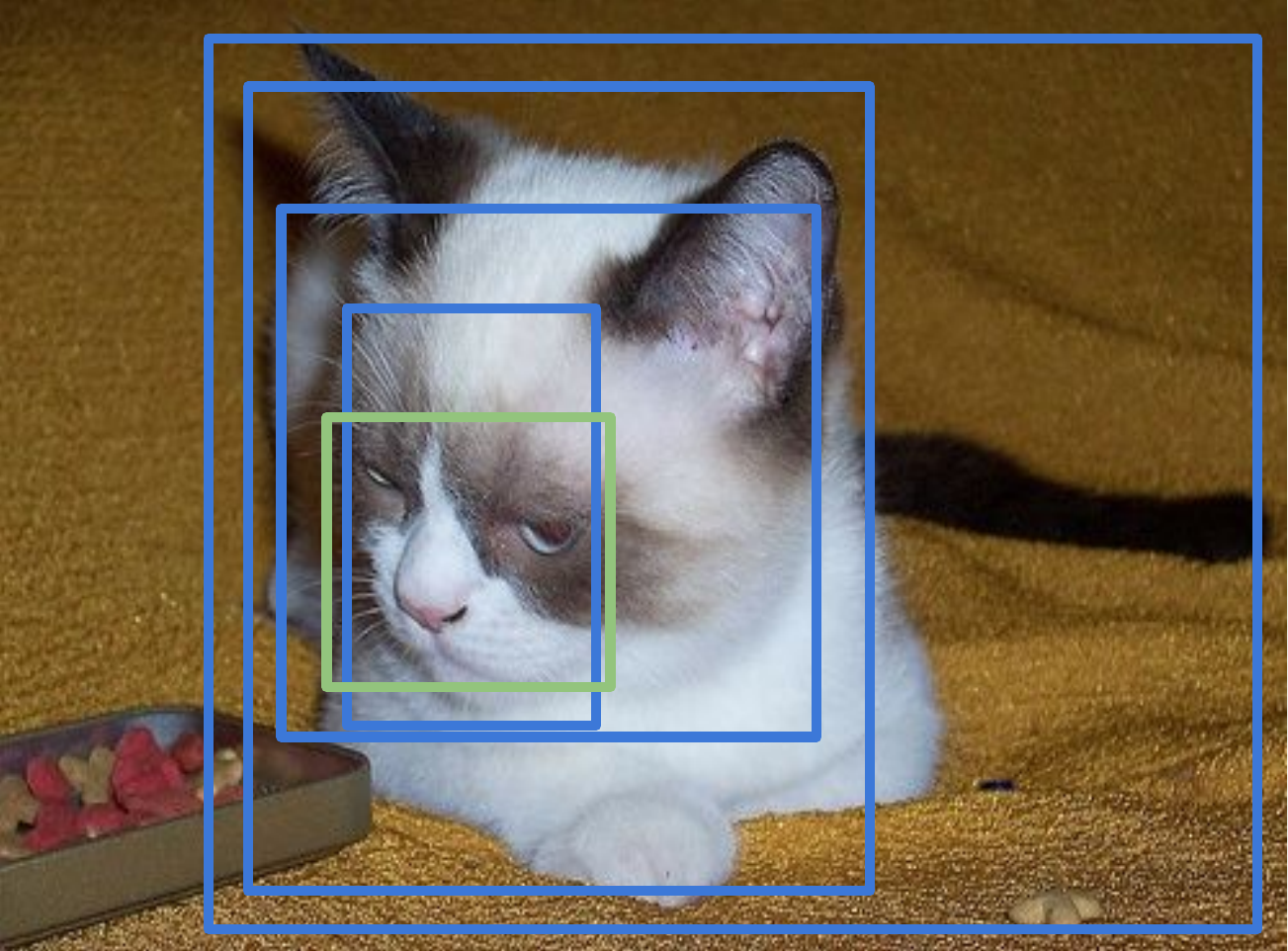}
    \caption*{$\lambda=0.01$}
\end{subfigure}
\caption{Effect of changing the IoU threshold $\lambda$ for instance selection. Green boxes are denote the supervision, blue boxes pass the threshold (selected) and red boxes fail (not selecetd).}
\label{fig_lambda_variation}
\end{figure}

\subsection{Classifier refinement agents}
\label{sec_refinement_agents}

To refine the outputs of the instance classifier, we use the online labeling and refinement strategy proposed by \cite{Tang2017_OIRC}. Here we refer to each $k^{th}$ refinement pass as $k^{th}$ refinement agent. 
In contrast with the instance classifier, each refinement agent outputs an additional dimension for background in its score vector $\mathbf{x}^{Rk}_j \in \mathbb{R}^{(C+1)\times 1}$, $k \in {1, 2, ...,K}$, where the $k$ is the index of the agent, $K$ is the total of agents, and the ${C + 1}^{th}$ dimension relates to the background.
The score vector from the instance classifier is represented here as  $\mathbf{x}^{R0}_j \in \mathbb{R}^{C\times 1}$, and is used to initialize the refinements. To obtain $\mathbf{x}^{Rk}_j$ for $k > 0$, the feature vector related to the proposals is passed through a single fc layer, and a softmax layer is applied over class dimension. 

Each agent needs some kind of supervision to learn how to separate the proposals related to the background from those related to ground-truth classes.  Thus, the supervision for agent $k$ is obtained from the previous agent $\mathbf{x}^{R(k-1)}$ and a supervision label vector is created for each proposal $j$ in the format $\mathbf{Y}^{k}_j = [y^k_{1j}, y^k_{2j}, \cdots, y^k_{(C+1),j}]^T \in  \mathbb{R}^{(C+1)\times 1}$. 
To build $\mathbf{Y}^{k}_j$, first the proposal with highest score is selected from the agent ${k- 1}^{th}$ supervision, sa given in Eq. \eqref{eq:best_instance}.
\begin{equation}
\label{eq:best_instance}
j^{k-1}_c = \argmax_r x^{R(k-1)}_{cr}.
\end{equation}
The highest score proposal is labeled as belonging to class $c$, i.e., $y^{k}_{cj^{k-1}_c} = 1$  and $y^{k}_{c'j^{k-1}_c} = 0$, $c'\neq c$.
Next, proposals with high overlap with $j^{k-1}_c$ are labeled as belonging to the same class of $j^{k-1}_c$, otherwise the adjacent proposals are labeled as background.  More precisely, this assignment is given by
\begin{equation}
\label{eq_selection}
{c^*}^k_{j}= \left\{\begin{matrix}
c, & \text{~~if~~} IoU(j^{k-1}_c, j^{k}_{cj}) \geq \lambda \\ 
C+1, &\text{otherwise}
\end{matrix}\right.,
\end{equation}
where $\lambda$ is the IoU threshold. We claim in this work that selecting a fixed value for $\lambda$ might not be the best choice, and present our dynamic threshold in Section \ref{smooth_supervision}. Each $y^k_{cj}$ is updated using ${c^*}^k_{j}$, that is, $y^k_{{c^*}^k_{j}j} = 1$. Meanwhile, if there is no object $c$ in the image, all values are set to zero, i.e.,  $y^k_{cj}=0$. 

Now that $y^k_{cj}$ is ready it can be used as supervision to train the $k^{th}$ refine agent using the loss function in Eq. \ref{eq_loss_ref_weigh}. 
\begin{equation}
\label{eq_loss_ref_weigh}
L^K_{agent} = - \frac{1}{|R|} \sum^{|R|}_{r=1}\sum^{C+1}_{c=1} w^k_r y^k_{cr} \log x^{Rk}_{cr}, 
\end{equation}
where $w^k_r$ is a weight term introduced to reduce noise during the supervision and is obtained as $w^k_r= x^{Rk-1}_{cj^{k-1}_c}$. More details can be found in~\cite{Tang2017_OIRC}.

\subsection{Knowledge distillation module}

The motivation behind cascading $K$ refinement agents in~\cite{Tang2017_OIRC} is that it allows the detector to gradually learn larger parts of objects, starting from the best instance only.  However, we can observe that the supervision generated by a $k^{th}$ agent will not be directly used by the ${k+2}^{th}$ agent. This happens because agent ${k+1}$ will learn with the supervision $k$ and will pass its own supervision to the next agent ${k+2}$. In other words, during the agent supervision process, some knowledge could be lost between the connections of the agents. We try to recover this information loss using our knowledge distillation module.  The distillation agent is a special kind of agent that learns using all the $K$ outputs as supervision. In reality, this agent only differs in the supervision part when compared with a standard refinement agent.

The distillation agent also outputs a score vector in the format $\mathbf{x}^{Dk}_j \in \mathbb{R}^{(C+1)\times 1}$. To obtain $\mathbf{x}^{Dk}_j$, the proposals-related feature vector is passed through a single fc layer, and a softmax layer is applied over the class dimension.

The supervision process of the distillation agent, instead of getting the supervision from a previous agent, uses all refinement agents outputs as supervision. More precisely, it is computed by averaging the outputs of the $K$ refinement agents outputs:
\begin{equation}
\label{eq:distillation_sup}
    \mathbf{x}^{D}_{cj} = \frac{1}{K} \sum^K_{k=1}  \mathbf{x}^{Rk}_{cj}.
\end{equation}

Using $\mathbf{x}^{D}_{cj}$ as the input to the supervision, the remaining process is similar to the described in section \ref{sec_refinement_agents} and the loss function $L_{destill}$ is the same as the weighted softmax loss in Eq. \eqref{eq_loss_ref_weigh}.

\subsection{Adaptive supervision aggregation function}
\label{smooth_supervision}

In~\cite{Tang2017_OIRC}, the authors experimentally chose $\lambda = 0.5$ as the proposal selection scheme in  Eq.~\eqref{eq_selection} to create the supervision matrices $w^k_r$ and $y^k_{cr}$. The interpretation of this value is that only boxes with $IoU > 0.5$ w.r.t. the best overall proposal are selected as belonging to the ground-truth class $c$. The problem with using a fixed value is that at the beginning of training, the instance selection module tends to select only small boxes as top score proposals, typically related to discriminant features of the objects (e.g., the face of a person or animal, as shown in Fig.~\ref{fig_lambda_variation}). As a consequence, only other small boxes will have $IoU > 0.5$ w.r.t. this box, and hence only small boxes will be considered as belonging to the class $c$. Figure~\ref{fig_lambda_variation} shows the effect of changing $\lambda$, where green denotes the best proposal, and blue the similar proposals according to the selected threshold.

\begin{figure}[t]
\begin{center}
  \includegraphics[width=0.75\linewidth]{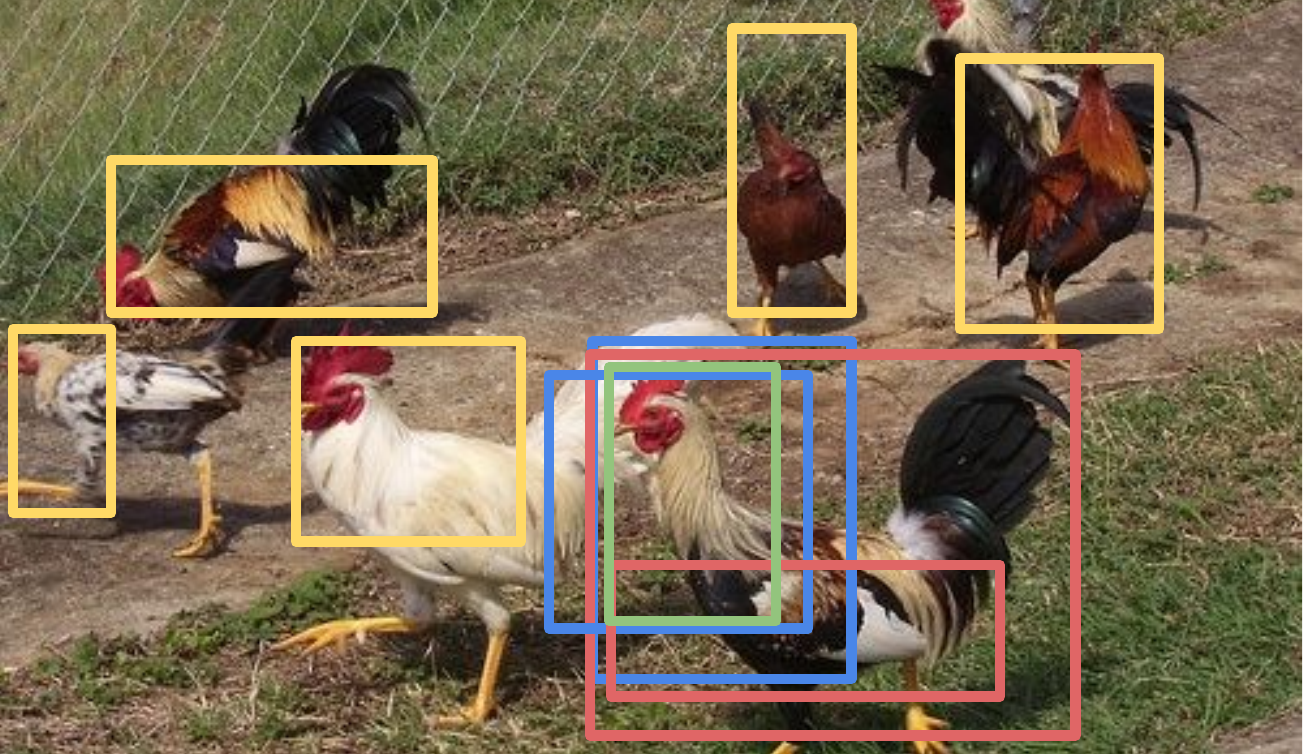}
\end{center}
\caption{
   A visual example of instance mining for ``chicken'' class, where the green box is the best instance. Boxes in blue present large IoU, in red present small (but not zero) IoU, and in yellow, the IoU is zero.       
}
\label{fig_the_trick}
\end{figure}
Although the goal of refinement agents is to gradually improve the detectors to find larger parts of objects, starting with a larger value for $\lambda$ causes each agent to highlight only small boxes in beginning of the process, and in some cases, the optimization will be stuck in small boxes during all training (especially for deformable objects). Relaxing $\lambda$ alleviates this issue, but it also tends to include proposals that are not related to the correct class.

Instead of using a fixed value for $\lambda$, we use an adaptive supervision function that changes  $\lambda$ during the training process. The function should be monotonically increasing, such that more candidates are aggregated in the beginning and less at the end. During our experiments, we evaluated a set of different adaptive supervision aggregation functions, and the best results were archived using the following function, also explored by C-MIL in a different context~\cite{Wan2019_CMIL}:
\begin{equation} 
\label{eq_lambda_function}
    \lambda = \frac{1}{2}\frac{\log (s+l_b) -\log l_b}{\log(S+l_b)-\log l_b},
\end{equation}
where $s$ is the current training step, $S$ is the total of training steeps, and $l_b$ defines the velocity that the curve grows.

\begin{figure}[htb]
\begin{center}
  \includegraphics[width=0.8\linewidth]{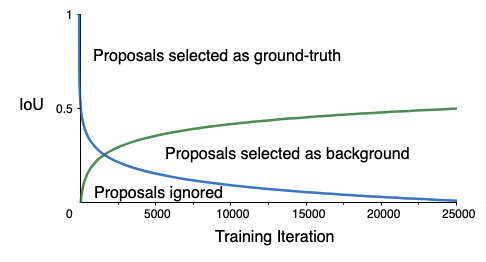}
\end{center}
\caption{
   A visual interpretation of the proposed adaptive supervision aggregation function. X-axis shows the iteration step number, and Y-axis shows the IoU with the box of the highest score.
}
\label{fig_smooth_function}
\end{figure}

Another deficiency of the supervision selection approach given by Eq. \eqref{eq_selection} is that  when more than one instance of a class is present in the image, it will obligatorily include all other instances as background in during the supervision
(since their $IoU$ with the best instance is small -- in general, null).  This is a bad decision, as we do not want to lower the scoring of these instances.  
In Fig.~\ref{fig_the_trick}, we present a visual example of this problem, considering the ``chicken'' class. In the figure, the rectangles are the candidate proposals, with the best one shown in green. Boxes shown in blue indicate proposals considered similar to the best one, according to Eq.~\eqref{eq_selection}, which leaves several proposals related to the chicken class (in yellow) marked as background, which is not desirable.

One solution to solve the penalization of other instances in the loss is to include the ``trick'' proposed by~\cite{Tang2018_RPN}, where a threshold value $\lambda_{ign}$ is used to ignore boxes with a low IoU w.r.t. $j^{k-1}_c$ in the loss. With the trick, all the instances of Fig.~\ref{fig_the_trick} in yellow would be ignored, and the ones in red would be marked as background.

In contrast to \cite{Tang2018_RPN}, where $\lambda_{ign}$ has a fixed value, we propose to use
an adaptive value similar to the scheme used for mining positive instances. Although the choice for $\lambda_{ign}$ could be independent from $\lambda$, we propose a ``complementary'' threshold selection scheme given by
\begin{equation} 
\label{eq_lambda_function_trick}
    \lambda_{ign} = \lambda_{max}-\lambda,
\end{equation} 
where $\lambda_{max}$ defines the starting point of the adaptive trick. 


Fig.~\ref{fig_smooth_function} presents the visual interpretation of $\lambda$ and $\lambda_{ign}$ during the supervision process. Thus, we can adapt Eq.~\eqref{eq_selection} to include the trick as is defined in Eq. \eqref{eq_selection_trick}, leading to
\begin{equation}
\label{eq_selection_trick}
{c^*}^k_{j}= \left\{\begin{matrix}
c,   & \text{~~if~~} IoU(j^{k-1}_c, j^{k}_{cj}) \geq \lambda, \\ 
C+1, & \text{~~if~~} IoU(j^{k-1}_c, j^{k}_{cj}) \geq \lambda_{ign} \\
-1 , &\text{otherwise}
\end{matrix}\right.,
\end{equation}
where $-1$ defines indices to be ignored in the agent loss functions.

\subsection{Final loss function}

The classification, refinement and distillations modules present individual loss functions. However, we train our model using a single loss that combine the individual loss functions given by
\begin{equation}
\label{eq:final_loss}
L = L_{class} + L_{distill} + \sum^K_{k=1} L^k_{agent}.
\end{equation}.

\section{Experiments}

Boosted-OICR was evaluated on the challenging PASCAL VOC 2007 and 2012 datasets \cite{everingham2010pascal}. Although the ground truth bounding box annotations are present in these datasets, we only use the (weak) classification annotations (presence or absence of a class in the given image). The performed evaluation is based on the two standard metrics in WSOD, that is, mean average precision (mAP)~\cite{everingham2010pascal} and correct localization (CorLoc)~\cite{deselaers2012weakly}. The former provides a measure of how well the detector adapts to all instances, while the latter indicates if the best detection is a good match. Both metrics utilizes PASCAL criteria of $IoU > 0.5$ between ground truths and predicted boxes.

\subsection{Implementation Details}

All experiments were performed using PyTorch 1.2~\cite{paszke2017automatic}\footnote{Source code available at: http://github.com/luiszeni/Boosted-OICR}. Our method uses VGG16~\cite{Simonyan2014} pre-trained on ImageNet~\cite{deng2009imagenet} as backbone. We replaced the last max-pooling layer by the SPP layer, and the last FC layer and softmax loss layer by the layers described in Section \ref{sec:method}. The new layers are initialized using Gaussian distributions with 0-mean and standard deviations 0.01. Biases are initialized to 0. The object proposals are extracted using Selective Search~\cite{Uijlings2013}. For data augmentation, the input images were re-sized into five scales $\{480, 576, 688, 864, 1200\}$ concerning the smallest image dimension. During training time, the scale of the image was randomly selected, and the image was randomly horizontal flipped, which is a standard approach among WSOD methods~\cite{Wan2019_CMIL, Tang2017_OIRC, Wan2019_MEL, Tang2018_PCL} and creates a total of ten augmented images. The learning process was done using the SGD algorithm with momentum 0.9, weight decay $5e^{-4}$, and batch size 2. We set $l_b=100$ and $\lambda_{max}=0.51$. The learning rate is set to 0.001 for the first 30K and 60K iterations and then decreases to 0.0001 in the following 20K and 30K iterations, respectively, for pascal VOC 2007 and 2012. During test time, all ten images are passed in the network, and the outputs are averaged. As an additional result, we also trained a supervised object detector by choosing top-scoring proposals as ground truth labels, as done in~\cite{Tang2017_OIRC, Tang2018_PCL, Wan2019_CMIL}. To make a fair comparison, we also trained a Fast RCNN (FRCNN)~\cite{Girshick2015} detection network using the five image scales. The supervision boxes are chosen by its score (larger than 0.3) and using non-maxima suppression (with 30\% IoU threshold).
\begin{table}[t]
\centering
\resizebox{0.8\linewidth}{!}{
\begin{tabular}{l|cccc|c}
\hline
ID      & K  & $\lambda$ & $\lambda_{ign}$  & distillation & mAP   \\ \hline
 1      & 3  & 0.5       & 0                & No           & 42.3         \\
 2      & 3  & adaptive       & 0                & No           & 41.6             \\
 3      & 3  & adaptive       & adaptive              & No           & 46.6             \\
 4      & 3  & adaptive       & adaptive              & Yes          & \textbf{49.7}           \\
 5      & 4  & adaptive       & adaptive              & No           & 48.1             \\
\hline
\end{tabular}
}
\caption{Ablation study performance (\%) on the VOC 2007.}
\label{tab_ablation_performance}
\end{table}

\begin{table*}[ht]
\centering
\resizebox{\textwidth}{!}{\begin{tabular}{l|lcccccccccccccccccccc|c}
\hline
Network                         & Method                             & aero & bike & bird & boat & bottle & bus  & car  & cat  & chair & cow  & table & dog  & horse & mbike & person & plant & sheep & sofa & train & tv   & mAP  \\ \hline
\multirow{10}{*}{VGG16}         & WSDDN \cite{Bilen2016}             & 46.4 & 58.3 & 35.5 & 25.9 & 14.0     & 66.7 & 53.0   & 39.2 & 8.9   & 41.8 & 26.6  & 38.6 & 44.7  & 59.0    & 10.8   & 17.3  & 40.7  & 49.6 & 56.9  & 50.8 & 39.2 \\
                                & OICR \cite{Tang2017_OIRC}          & 58.0   & 62.4 & 31.1 & 19.4 & 13.0     & 65.1 & 62.2 & 28.4 & 24.8  & 44.7 & 30.6  & 25.3 & 37.8  & 65.5  & 15.7   & 24.1  & 41.7  & 46.9 & 64.3  & 62.6 & 42.0   \\
                                & WCCN \cite{Diba2017}               & 49.5 & 60.6 & 38.6 & 29.2 & 16.2   & 70.8 & 56.9 & 42.5 & 10.9  & 44.1 & 29.9  & 42.2 & 47.9  & 64.1  & 13.8   & 23.5  & 45.9  & 54.1 & 60.8  & 54.5 & 42.8 \\
                                & TS2C \cite{Wei2018}                & 59.3 & 57.5 & 43.7 & 27.3 & 13.5   & 63.9 & 61.7 & 59.9 & 24.1  & 46.9 & 36.7  & 45.6 & 39.9  & 62.6  & 10.3   & 23.6  & 41.7  & 52.4 & 58.7  & 56.6 & 44.3 \\
                                & WeakRPN \cite{Tang2018}            & 57.9 & \textbf{70.5} & 37.8 & 5.7  & 21.0     & 66.1 & 69.2 & 59.4 & 3.4   & 57.1 & \textbf{57.3}  & 35.2 & \textbf{64.2}  & 68.6  & \textbf{32.8}   & \textbf{28.6}  & 50.8  & 49.5 & 41.1  & 30.0   & 45.3 \\
                                & PCL \cite{Tang2018_PCL}            & 54.4 & 69.0   & 39.3 & 19.2 & 15.7   & 62.9 & 64.4 & 30.0   & \textbf{25.1}  & 52.5 & 44.4  & 19.6 & 39.3  & 67.7  & 17.8   & 22.9  & 46.6  & 57.5 & 58.6  & 63   & 43.5 \\
                                & MELM \cite{Wan2019_MEL}            & 55.6 & 66.9 & 34.2 & 29.1 & 16.4   & 68.8 & 68.1 & 43.0   & 25.0    & \textbf{65.6} & 45.3  & 53.2 & 49.6  & 68.6  & 2.0      & 25.4  & \textbf{52.5}  & 56.8 & 62.1  & 57.1 & 47.3 \\
                                & C-MIL \cite{Wan2019_CMIL}          & 62.5 & 58.4 & 49.5 & \textbf{32.1} & 19.8   & 70.5 & 66.1 & \textbf{63.4} & 20.0    & 60.5 & 52.9  & 53.5 & 57.4  & 68.9  & 8.4    & 24.6  & 51.8  & \textbf{58.7} & 66.7  & 63.5 & \textbf{50.5} \\
                                \cline{2-23} 
                                & Ours & \textbf{68.6} & 62.4 & \textbf{55.5} & 27.2 & \textbf{21.4} & \textbf{71.1} & \textbf{71.6} & 56.7 & 24.7 & 60.3 & 47.4 & \textbf{56.1} & 46.4 & \textbf{69.2} & 2.7 & 22.9 & 41.5 & 47.7 & \textbf{71.1} & \textbf{69.8} & 49.7 \\
                                \hline
\multirow{6}{*}{\shortstack[l]{FRCNN\\Re-train}} & OICR \cite{Tang2017_OIRC}          & 65.5 & 67.2 & 47.2 & 21.6 & 22.1   & 68.0   & 68.5 & 35.9 & 5.7   & 63.1 & 49.5  & 30.3 & 64.7  & 66.1  & 13.0     & 25.6  & 50.0    & 57.1 & 60.2  & 59.0   & 47.0   \\
                                & TS2C \cite{Wei2018}                & -    & -    & -    & -    & -      & -    & -    & -    & -     & -    & -     & -    & -     & -     & -      & -     & -     & -    & -     & -    & 48.0   \\
                                & PCL \cite{Tang2018_PCL}            & 63.2 & \textbf{69.9} & 47.9 & 22.6 & 27.3   & 71.0   & 69.1 & 49.6 & 12.0    & 60.1 & 51.5  & 37.3 & 63.3  & 63.9  & 15.8   & 23.6  & 48.8  & 55.3 & 61.2  & 62.1 & 48.8 \\
                                & WeakRPN \cite{Tang2018}            & 63.0   & 69.7 & 40.8 & 11.6 & \textbf{27.7}   & 70.5 & \textbf{74.1} & 58.5 & 10.0    & \textbf{66.7} & \textbf{60.6}  & 34.7 & \textbf{75.7}  & \textbf{70.3}  & \textbf{25.7}   & \textbf{26.5}  & \textbf{55.4}  & 56.4 & 55.5  & 54.9 & 50.4 \\
                                & C-MIL \cite{Wan2019_CMIL}          & 61.8 & 60.9 & \textbf{56.2} & 28.9 & 18.9   & 68.2 & 69.6 & \textbf{71.4} & 18.5  & 64.3 & 57.2  & 66.9 & 65.9  & 65.7  & 13.8   & 22.9  & 54.1  & \textbf{61.9} & 68.2  & \textbf{66.1} & \textbf{53.1} \\ \cline{2-23} 
                                &

Ours & \textbf{65.8} & 58.6 & 55.0 & \textbf{32.4} & 19.5 & \textbf{74.2} & 71.4 & 70.9 & \textbf{19.2} & 54.8 & 46.2 & \textbf{67.5} & 57.0 & 65.6 & 1.4 & 16.7 & 40.4 & 53.0 & \textbf{69.5} & 61.1 & 50.0 \\ \hline
\end{tabular}}
\caption{Detection performance (\%) on the VOC 2007 test set. Comparison to the state-of-the-arts.} 
\label{tab_voc_2007_detection}
\end{table*}
\begin{table*}[ht]
\centering
\resizebox{\textwidth}{!}{\begin{tabular}{clcccccccccccccccccccc|c}
\hline
\multicolumn{1}{l}{Network}                 & Method                  & aero & bike & bird & boat & bottle & bus  & car  & cat  & chair & cow  & table & dog  & horse & mbike & person & plant & sheep & sofa & train & tv   & mAP  \\ \hline
\multicolumn{1}{c|}{\multirow{9}{*}{VGG16}} & WSDDN \cite{Bilen2016}  & 65.1 & 58.8 & 58.5 & 33.1 & 39.8   & 68.3 & 60.2 & 59.6 & 34.8  & 64.5 & 30.5  & 43.0   & 56.8  & 82.4  & 25.5   & 41.6  & 61.5  & 55.9 & 65.9  & 63.7 & 53.5 \\
\multicolumn{1}{c|}{}                       & OICR \cite{Tang2017_OIRC}   & 81.7 & 80.4 & 48.7 & 49.5 & 32.8   & 81.7 & 85.4 & 40.1 & 40.6  & 79.5 & 35.7  & 33.7 & 60.5  & 88.8  & 21.8   & 57.9  & 76.3  & 59.9 & 75.3  & 81.4 & 60.6 \\
\multicolumn{1}{c|}{}                       & WCCN \cite{Diba2017}    & 83.9 & 72.8 & 64.5 & 44.1 & 40.1   & 65.7 & 82.5 & 58.9 & 33.7  & 72.5 & 25.6  & 53.7 & 67.4  & 77.4  & 26.8   & 49.1  & 68.1  & 27.9 & 64.5  & 55.7 & 56.7 \\
\multicolumn{1}{c|}{}                       & TS2C \cite{Wei2018}     & 84.2 & 74.1 & 61.3 & 52.1 & 32.1   & 76.7 & 82.9 & 66.6 & 42.3  & 70.6 & 39.5  & 57.0   & 61.2  & 88.4  & 9.3    & 54.6  & 72.2  & 60.0   & 65.0    & 70.3 & 61.0   \\
\multicolumn{1}{c|}{}                       & WeakRPN \cite{Tang2018} & 77.5 & 81.2 & 55.3 & 19.7 & 44.3   & 80.2 & 86.6 & \textbf{69.5} & 10.1  & \textbf{87.7} & \textbf{68.4}  & 52.1 & \textbf{84.4}  & \textbf{91.6}  & \textbf{57.4}   & \textbf{63.4}  & \textbf{77.3}  & 58.1 & 57.0    & 53.8 & 63.8 \\
\multicolumn{1}{c|}{}                       & PCL \cite{Tang2018_PCL}    & 79.6 & \textbf{85.5} & 62.2 & 47.9 & 37.0     & \textbf{83.8} & 83.4 & 43.0   & 38.3  & 80.1 & 50.6  & 30.9 & 57.8  & 90.8  & 27.0     & 58.2  & 75.3  & \textbf{68.5} & 75.7  & 78.9 & 62.7 \\
\multicolumn{1}{c|}{}                       & MELM \cite{Wan2019_MEL}     & -    & -    & -    & -    & -      & -    & -    & -    & -     & -    & -     & -    & -     & -     & -      & -     & -     & -    & -     & -    & 61.4 \\
\multicolumn{1}{c|}{}                       & C-MIL \cite{Wan2019_CMIL}   & -    & -    & -    & -    & -      & -    & -    & -    & -     & -    & -     & -    & -     & -     & -      & -     & -     & -    & -     & -    & 65.0 \\ \cline{2-23} 
\multicolumn{1}{c|}{}                       & Ours & \textbf{86.7} & 73.3 & \textbf{72.4} & \textbf{55.3} & \textbf{46.9} & 83.2 & \textbf{87.5} & 64.5 & \textbf{44.6} & 76.7 & 46.4 & \textbf{70.9} & 67.0 & 88.0 & 9.6 & 56.4 & 69.1 & 52.4 & \textbf{79.8} & \textbf{82.8} & \textbf{65.7}\\ \hline
\end{tabular}}
\caption{Localization performance (\%) on the VOC 2007 trainval set. Comparison to the state-of-the-arts.} 
\label{tab_voc_2007_corloc}
\end{table*}
\subsection{Ablation experiments}

We conduct some ablation experiments to illustrate the effectiveness of the proposed improvements over the baseline method OICR~\cite{Tang2017_OIRC}.


\begin{table}[t]
\centering
\resizebox{.5\linewidth}{!}{
\begin{tabular}{l|c|c}
\hline
Method                             & mAP  & Corloc \\ \hline
WCCN \cite{Diba2017}               & 37.9 & -      \\
OICR \cite{Tang2017_OIRC}          & 37.9 & 62.1   \\
TS2C \cite{Wei2018}                & 40   & 64.4   \\
WeakRPN \cite{Tang2018}            & 40.8 & 64.9   \\
PCL \cite{Tang2018_PCL}            & 40.6 & 63.2   \\
MELM \cite{Wan2019_MEL}            & 42.4 & -      \\
C-MIL \cite{Wan2019_CMIL}          & \textbf{46.6} & \textbf{67.4}   \\ \hline
Ours &  *   & 66.3 \\ \hline
\end{tabular}
}
\caption{Detection (test set) and localization (trainval set) performance (\%) on the VOC 2012 dataset using VGG16. }  
\label{tab_voc_2012}
\end{table}

\begin{figure*}[t]
\begin{center}
  \includegraphics[width=0.98\linewidth]{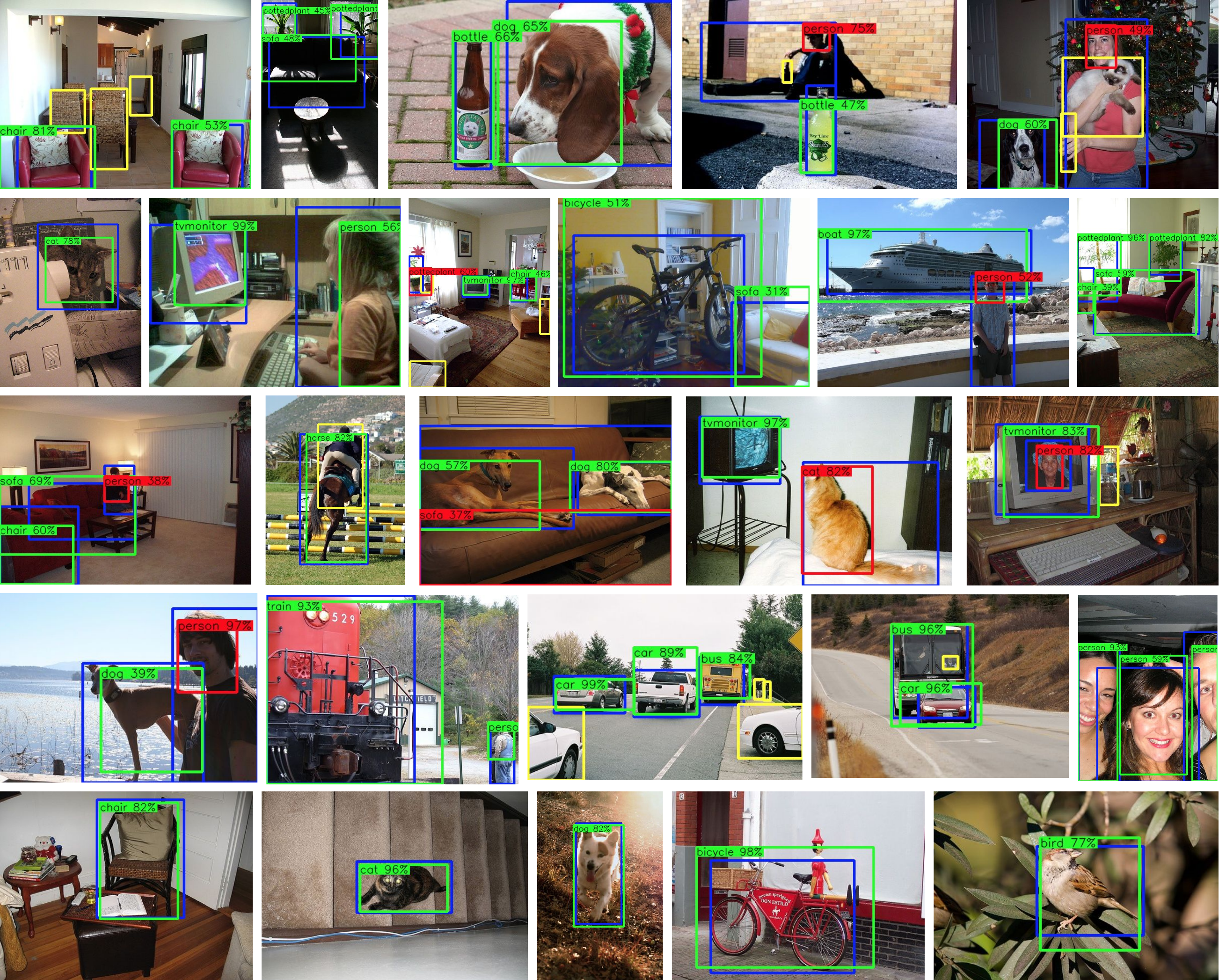}
\end{center}
\caption{
    Detection examples for Pascal VOC 2007 dataset. Blue rectangles are ground-truth boxes that have at least one detection with $IoU>0$, and yellow ones are ground-truth with no detection intersection. Green boxes are correct detections ($IoU>0.5$ with ground truth), and red boxes are wrong detections. The label in each detection box is the class label and confidence score of the detection. 
}
\label{fig:results}
\end{figure*}

We first study the impact of using the adaptive supervision aggregation function instead of fixed IoU thresholds for proposal mining. We display the different scenarios in Table \ref{tab_ablation_performance}. The experiment with ID$=1$ presents the results using the standard OICR pipeline.
In the experiment ID$=2$ we replace the fixed $\lambda$ value by the proposed adaptive aggregation function defined in Eq.~\eqref{eq_lambda_function}, in this experiment all boxes with $IoU < \lambda$ are considered as background. As the experiment suggests, using the adaptive supervision aggregation function alone without the adaptive trick makes the results worse than the OICR's baseline.
However, adding the adaptive trick (experiment ID=$3$) leads to  an improvement of $4.3\%$ in the final mAP, suggesting that using our adaptive supervision aggregation function can boost the OICR detection mAP significantly.



We also evaluated the effect of including the distillation refinement module. In fact, one could argue that using such a module could produce the same result as cascading one more refinement agent. To show the difference, we tested our method using $K=4$ (and no distillation) vs. $K=3$ with distillation, and results with distillation were considerably better (see experiments ID$=4$ vs. ID$=5$ in Table \ref{tab_ablation_performance}).  As we can see, adding the knowledge distillation improves the results in $1.6\%$ mAP more than adding an extra refinement agent. We select the model utilized in the experiment ID=$4$ as default to the next experiments. 


\subsection{Comparison with state-of-the-art} 

We compare our results with other state-of-the-art (SOTA) methods in the Pascal VOC 2007 and 2012 datasets. 
Table~\ref{tab_voc_2007_detection} shows a comparison of detection performance of our method and SOTA in the Pascal VOC 2007 test set. It can be seen that Boosted-OICR improves the original OICR paper~\cite{Tang2017_OIRC} in $7.7\%$ mAP and outperformed other approaches such as
WCCN~\cite{Diba2017} ($6.9\%$), TS2C~\cite{Wei2018} ($5.4\%$), WeakRPN~\cite{Tang2018} ($4.4\%$), PCL~\cite{Tang2018_PCL} ($6.2\%$), and MELM~\cite{Wan2019_MEL} ($2.4\%$). Boosted-OICR was only inferior to C-MIL \cite{Wan2019_CMIL} by a small value ($0.8\%$ mAP). However, Boosted-OICR presented the highest AP results in 9 of the total 20 classes (\texttt{aeroplane},  \texttt{bird}, \texttt{bottle}, \texttt{bus}, \texttt{car}, \texttt{dog}, \texttt{motorbike}, \texttt{train} and \texttt{tv}). Figure~\ref{fig:results} presents some results generated by our WSOD method. We also re-trained an Fast-RCNN detector using the learned pseudo objects as ground-truth, and achieved $50\%$ mAP, as shown in Table~\ref{tab_voc_2007_detection}, which improved our method by $0.3\%$. 

Table~\ref{tab_voc_2007_corloc} presents a comparison in localization performance of our method and SOTA in the Pascal VOC 2007 train-val set. Boosted-OICR outperformed
OICR~\cite{Tang2017_OIRC} ($5.1\%$),
WCCN~\cite{Diba2017} ($9.0\%$), TS2C~\cite{Wei2018} ($4.7\%$), WeakRPN~\cite{Tang2018} ($1.9\%$), PCL~\cite{Tang2018_PCL} ($3.0\%$), MELM~\cite{Wan2019_MEL} ($4.3\%$), and C-MIL~\cite{Wan2019_CMIL} ($0.7\%$). The better corloc result of our method in comparison with C-MIL suggests that C-MIL is just a little better dealing with images with more than one instance (which impacts the final detection mAP).
We also compare the localization performance of our method in pascal VOC 2012\footnote{\label{voc_footnote}
    We submitted our results for VOC 2012 to the evaluation server, but still did not get the feedback. The anonymous submission link is http://host.robots.ox.ac.uk:8080/anonymous/E7JSMD.html}. 
in Table~\ref{tab_voc_2012}.  Boosted-OICR presents a competitive corloc in VOC 2012 outperforming OICR~\cite{Tang2017_OIRC} ($4.2\%$),
TS2C~\cite{Wei2018} ($1.9\%$), WeakRPN~\cite{Tang2018} ($1.4\%$) and PCL~\cite{Tang2018_PCL} ($3.1\%$), being inferior to C-MIL \cite{Wan2019_CMIL} by  $1.1\%$ mAP.


\section{Conclusions}
In this paper, we propose two improvements to boost the online instance classifier refinement. First, we propose a knowledge distillation methodology that extracts extra knowledge from the refinement agents. Second, we propose an adaptive supervision aggregation function that improves the way that each refinement agent learns to separate class-related instances,  background instances, and which instances ignore. 
Both contributions were built using OICR as a baseline approach, and the proposed contributions were able to provide a $7.4$ mAP boost over the OICR baseline method.
Boosted-OICR presents competitive SOTA results on Pascal VOC 2007 dataset, being inferior only to~\cite{Wan2019_CMIL} by a small margin ($0.8\%$ mAP). Also, Boosted-OICR presents the highest AP results in 9 of the 20 classes, such as \texttt{airplane}, \texttt{bird}, \texttt{bottle}, and \texttt{train}. Although  Boosted-OICR has the best performance in these classes, it fails in deformable objects such as \texttt{person} class. In fact, the \texttt{person} class is very challenging, since the GT annotations might contain only the face or upper body (when there are occlusions), or the whole body.

In the future, we intend to explore improvements that make WSOD methods to not focus on the most discriminated part of deformable objects such as the human face.
We further plan to explore mid-layers of the network and class activation maps to create object proposals as an alternative to the selective search module.

\section*{Acknowledgments}
The authors would like to thank Brazilian funding agencies CNPq and CAPES (Finance Code 001), as well as NVIDIA Corporation for the donation of a Titan Xp Pascal GPU used for this research.

{\small
\bibliographystyle{ieee_fullname}
\bibliography{egbib}
}

\end{document}